\documentclass[lettersize,journal]{IEEEtran}
\usepackage{algorithm}
\usepackage{array}
\usepackage[caption=false,font=normalsize,labelfont=sf,textfont=sf]{subfig}
\usepackage{stfloats}
\usepackage{url}
\usepackage{verbatim}
\usepackage{cite}
\usepackage[hidelinks]{hyperref}
\usepackage{colortbl}
\definecolor{tablegray}{gray}{0.9} 
\definecolor{textgray}{gray}{0.5}
\usepackage{amsmath,amssymb,amsfonts}
\usepackage{dsfont}
\usepackage{algpseudocode}
\usepackage{makecell}
\usepackage{multirow}

\usepackage{graphicx}
\usepackage{textcomp}

\begin{document}

\title{OphBiWSSD: Scaling Temporal Action Localization in Ophthalmic Surgeries with Bidirectional Weight-tied State Space Duality}
\author{
Yang Liu, Qionghong Ma, JOONGWON CHAE, Lihui Luo, Yibing Shen, Yulin Zhuo, Yingting Zhu, Jiashu Chang, Xiaoyun Zhong, Dongmei Yu, Peter E. Lobie, Peiwu Qin, Chengming Yang
\thanks{This work was supported in part by the Joint TCM Science \& Technology Projects of National Demonstration Zones for Comprehensive TCM Reform, GZY-KJS-ZJ-2026-032; National Natural Science Foundation of China (No.22306048, 62473005, U24A20761); Joint Fund of Zhejiang Provincial Natural Science Foundation of China, LLSQN26H090003; Sichuan Provincial Natural Science Foundation of China, 2026NSFSC1501; Medical and Health Science Program of Zhejiang Province, 2025HY1542; Zhejiang Province Post-Doctoral Research Project Selected Funding, Z12024150; Shenzhen Science, and Technology Program (No.JCYJ20250604145715020, JCYJ20230807113017035, JCYJ20240813112016022); and Shenzhen High-Level and Urgently Needed TalentStartup Funding. \textit{(Corresponding author: Peiwu Qin, Chengming Yang)}}
\thanks{Yang Liu, JOONGWON CHAE, Lihui Luo, Yibing Shen, Xiaoyun Zhong, and Peter E. Lobie are with the
Institute of Biopharmaceutics and Health Engineering, Tsinghua Shenzhen International Graduate School, Shenzhen 518055, China (e-mail: lyang22@mails.tsinghua.edu.cn; cai-zy24@mails.tsinghua.edu.cn; luolh23@mails.tsinghua.edu.cn; shenyb25@mails.tsinghua.edu.cn; heulwenchung@gmail.com; pelobie@sz.tsinghua.edu.cn).}
\thanks{Qionghong Ma and Dongmei Yu are with Department of Ophthalmology, The Fifth Affiliated Hospital of Wenzhou Medical University, Wenzhou Medical University, Lishui 323000, China (e-mail: 22118515@zju.edu.cn; ydmei1979@163.com).}
\thanks{Yulin Zhuo is with School of Optometry, The Hong Kong Polytechnic University (e-mail: yulinzhuoedu@outlook.com).}
\thanks{Yingting Zhu is with State Key Laboratory of Ophthalmology, Zhongshan Ophthalmic Center, Sun Yat-sen University, Guangdong Provincial Key Laboratory of Ophthalmology Visual Science, Guangdong Provincial Clinical Research Center for Ocular Diseases, Guangzhou 510060, China (e-mail: zhuyt35@mail.sysu.edu.cn).} 
\thanks{Jiashu Chang is with Faculty of Science and Technology, Beijing Normal-Hong Kong Baptist University, Zhuhai 519087, China (e-mail: s230034004@mail.uic.edu.cn).}
\thanks{Peiwu Qin is with Chinese Medicine Guangdong Laboratory/Hengqin Laboratory, Hengqin, Guangdong 519031, China (e-mail: pwqin1979@gmail.com).}
\thanks{Chengming Yang is with Southern University of Science and Technology Hospital, Shenzhen 518055, China (e-mail: xluckyy\_scholar@163.com).}
}

\markboth{Journal of \LaTeX\ Class Files,~Vol.~14, No.~8, August~2021}%
{Shell \MakeLowercase{\textit{et al.}}: A Sample Article Using IEEEtran.cls for IEEE Journals}


\maketitle

\begin{abstract}
High-frequency surgical maneuvers in ophthalmology necessitate high-fidelity temporal modeling, yet characterizing long-range procedural dependencies remains computationally prohibitive for attention-based architectures. Existing models often require aggressive temporal downsampling, which compromises the detection of fine-grained action boundaries and instrument-tissue interactions. To address these scalability constraints, we present OphBiWSSD, a framework that reformulates surgical temporal action localization leveraging Bidirectional State Space Duality. By employing a weight-tied selective scan mechanism that incorporates both preceding and succeeding surgical contexts, our approach facilitates the global synthesis of non-causal temporal cues with linear complexity. This streamlined architecture is well-suited to capture the bidirectional dependencies present in ophthalmic workflows, effectively bridging the gap between local boundary precision and long-range procedural context without incurring the quadratic memory overhead of traditional Transformers. Extensive experiments on the OphNet benchmark demonstrate that OphBiWSSD achieves state-of-the-art temporal localization performance, with mean Average Precisions of 44.42\% on phases and 43.08\% on operations, surpassing the baselines by 6.80\% and 6.66\%, respectively. Empirical validation indicates that our approach ensures precise temporal localization and offers a computationally viable pathway for deploying surgical intelligence systems in clinical environments. The code is publicly available at \href{https://github.com/yo3nglau/OphBiWSSD}{https://github.com/yo3nglau/OphBiWSSD}.
\end{abstract}


\section{Introduction}

The emergence of Surgical Data Science (SDS)~\cite{maier2017surgical,maier2022surgical} has catalyzed a paradigm shift in intraoperative safety, objective skill assessment, and automated workflow analysis. Central to these advancements is Temporal Action Localization (TAL), a task that simultaneously identifies surgical action categories and regresses their precise temporal boundaries~\cite{zhao2017temporal,lea2017temporal}. In ophthalmic surgery, characterized by microscopic operative fields and high-frequency maneuvers~\cite{schoeffmann2018cataract,zisimopoulos2018deepphase}, the ability to accurately segment procedural milestones and atomic tool-tissue interactions is fundamental to providing context-aware intraoperative assistance and postoperative feedback~\cite{hashimoto2018artificial}.

Contemporary TAL architectures primarily rely on self-attention mechanisms~\cite{vaswani2017attention}. Existing models encounter significant challenges in balancing temporal resolution with computational feasibility. In ophthalmic procedures, high-frame-rate video acquisition is essential for accurately delineating sub-second gesture transitions. The resulting sequence length poses a substantial computational barrier, as the complexity of self-attention scales quadratically with the input length. Although established models such as ActionFormer~\cite{zhang2022actionformer} and TriDet~\cite{shi2023tridet} demonstrate strong performance, aggressive temporal decimation compromises the precise localization of dense surgical boundaries. This limitation underscores the need for architectures that preserve full temporal resolution without incurring the excessive memory and computational demands of standard Transformer-based designs.

State Space Models (SSMs)~\cite{gu2021combining,gu2022efficiently} have emerged as computationally viable alternatives to the attention paradigm. The Mamba~\cite{gu2024mamba} framework introduces a selective scan mechanism that enables content-based reasoning with linear complexity in terms of sequence length. By leveraging matrix-recurrent duality and hardware-aware parallel scans, SSMs facilitate efficient modeling of long temporal horizons that were previously inaccessible to quadratic architectures. Although originally developed for sequential data such as natural language, the adaptation of semiseparable matrices and hardware-optimized kernels in the State Space Duality (SSD)~\cite{dao2024transformers} framework offers a robust approach for processing dense, high-resolution sequences typical of ophthalmic surgery. Despite the efficiency of standard SSMs, their formulations are inherently causal, as they derive the latent state at any given point exclusively from preceding information. This unidirectional constraint is suboptimal for surgical TAL, as precise action identification often requires observing subsequent events. For example, identifying a corneal incision may depend on recognizing a subsequent tool withdrawal or specific tissue responses to resolve temporal ambiguity.

To address these limitations, we propose OphBiWSSD, a framework that reformulates surgical TAL through Bidirectional Weight-tied State Space Duality. Our approach achieves linear computational complexity while facilitating the global, non-causal synthesis of temporal features. By integrating the associative properties of Transformers with the computational efficiency of SSMs, OphBiWSSD masterfully bridges the gap between local boundary precision and long-range procedural context. Our principal contributions are summarized as follows:

(\textbf{i}) A Bidirectional Selective Scan mechanism that adeptly captures anti-causal surgical dependencies without quadratic memory overhead.

(\textbf{ii}) A Weight-tied State Space Duality formulation that functions as a structural regularizer, enforcing direction-invariant feature learning across both forward and backward scanning paths.

(\textbf{iii}) A new state-of-the-art performance on the OphNet benchmark, achieving mean Average Precisions of 44.42\% on phases and 43.08\% on operations, demonstrating a significant improvement over previous attention-based architectures.

\section{Related Works}

\subsection{Temporal Action Localization}

Temporal Action Localization (TAL) in long-form video involves simultaneously classifying action categories and regressing to precise temporal boundaries. Traditional paradigms~\cite{zhao2017temporal,xu2017r} often relied on multi-stage pipelines that involved proposal generation followed by boundary refinement. While these decoupled approaches provided modularity, they often suffered from suboptimal error propagation across stages and high inference latency~\cite{shou2017cdc}. Recently, anchor-free, single-stage architectures~\cite{lin2021learning} have gained prominence. These models streamline the localization pipeline by directly regressing boundary locations from feature maps, thereby enabling end-to-end optimization and maintaining higher temporal resolution. ActionFormer~\cite{zhang2022actionformer} utilized a multi-scale Transformer-based encoder to leverage the global receptive field of self-attention for boundary prediction. TriDet~\cite{shi2023tridet} further advanced this by introducing a Trident head to better model boundary confidence via relative displacement. Recent iterations such as DyFADet~\cite{yang2024dyfadet} and CLTDR-GMG~\cite{li2025temporal} have explored dynamic feature aggregation and task decoupling for temporal refinement. However, these architectures predominantly rely on self-attention mechanisms, which incur quadratic computational complexity with respect to the sequence length. This scalability bottleneck often necessitates substantial temporal pooling, which can obscure the fine-grained transitions critical for high-fidelity surgical analysis.

\subsection{Surgical Workflow Analysis}

Surgical Data Science (SDS) plays a critical role in enhancing intraoperative safety and facilitating objective skill assessment through automated video analysis. The field has evolved from general surgical phase recognition to more granular tasks, including temporal action recognition and localization, which demand both gesture identification and precise temporal boundary determination. Although substantial progress~\cite{perez2024must,liu2025lovit} has been achieved in laparoscopic and endoscopic surgery benchmarks~\cite{nwoye2022rendezvous,wagner2023comparative}, ophthalmic surgery remains challenging due to rapid maneuvers and subtle visual cues inherent in instrument-tissue interactions~\cite{schoeffmann2018cataract,ghamsarian2024cataract}. The OphNet~\cite{hu2024ophnet} dataset introduces a comprehensive, hierarchical framework that addresses these challenges and establishes essential baselines using ActionFormer~\cite{zhang2022actionformer} and TriDet~\cite{shi2023tridet}, thereby advancing clinical surgical intelligence.

\subsection{State Space Models}

State Space Models (SSMs)~\cite{gu2021combining,gu2022efficiently}, particularly the Mamba~\cite{gu2024mamba} architecture, have emerged as a computationally efficient alternative to Transformers for sequence modeling. By utilizing a selective scan mechanism, Mamba achieves linear scaling while maintaining a content-aware receptive field. This innovation has been extended to the vision domain through Vision Mamba~\cite{zhu2024vision} and VMamba~\cite{liu2024vmamba}, which introduce bidirectional or cross-scan modules to handle non-causal spatial data. In video understanding, VideoMamba~\cite{li2024videomamba} and the Video Mamba Suite~\cite{chen2026video} have demonstrated that SSMs can effectively model long-range temporal dependencies with significantly lower memory overhead than self-attention. The recent evolution toward Mamba-2~\cite{dao2024transformers} introduces SSD, which establishes a theoretical bridge between structured SSMs and semiseparable matrices, enabling faster training through well-optimized matrix multiplications.

\section{Methodology}

\subsection{Preliminaries}

The foundational SSMs map a one-dimensional input sequence $x(t) \in \mathbb{R}$ to an output $y(t) \in \mathbb{R}$ through an intermediate latent state $h(t) \in \mathbb{R}^N$. This system is governed by a first-order linear ordinary differential equation (ODE):
\begin{equation}
\dot{h}(t) = \mathbf{A}h(t) + \mathbf{B}x(t), \quad y(t) = \mathbf{C}h(t)
\end{equation}

where $\mathbf{A} \in \mathbb{R}^{N \times N}$ represents the state transition matrix, $\mathbf{B} \in \mathbb{R}^{N \times 1}$ is the input matrix, and $\mathbf{C} \in \mathbb{R}^{1 \times N}$ is the projection matrix. To transform the continuous-time representation into a discrete-time framework suitable for digital sequence processing, the system parameters in (1) undergo a discretization process. Applying the Zero-Order Hold (ZOH) method with a discretization step size $\Delta$ yields the discrete-time parameters $(\mathbf{\bar{A}}, \mathbf{\bar{B}})$:
\begin{equation}
\mathbf{\bar{A}} = \exp(\Delta \mathbf{A}), \quad \mathbf{\bar{B}} = (\Delta \mathbf{A})^{-1}(\exp(\Delta \mathbf{A}) - \mathbf{I}) \cdot \Delta \mathbf{B}
\end{equation}

The discrete recurrence relation is subsequently defined as:
\begin{equation}
h_k = \mathbf{\bar{A}}h_{k-1} + \mathbf{\bar{B}}x_k, \quad y_k = \mathbf{C}h_k
\end{equation}

While early Structured SSMs (S4)~\cite{gu2022efficiently} utilized Linear Time-Invariant (LTI) kernels, the Mamba architecture introduces the Selective Scan Mechanism (S6)~\cite{gu2024mamba}. In this formulation, $(\mathbf{B}, \mathbf{C}, \Delta)$ are defined as dynamic functions of the input $x_k$, enabling the model to selectively propagate or attenuate information across the temporal axis $k$.

Structured SSD~\cite{dao2024transformers} further refines this mechanism by constraining $\mathbf{A}$ to a scalar-times-identity structure, where $\mathbf{A} = a\mathbf{I}$. This constraint facilitates a theoretical duality between the recurrent view in (3) and a matrix-multiplication view. Specifically, the transformation can be expressed as a semiseparable matrix multiplication:
\begin{equation}
\mathbf{y} = \mathbf{M} \mathbf{x}, \quad \text{where } \mathbf{M}_{jk} = \mathbf{C}_j \left( \prod_{i=k+1}^j \mathbf{\bar{A}}_i \right) \mathbf{\bar{B}}_k
\end{equation}

This duality permits the use of hardware-efficient kernels during training, mirroring the parallelization of Transformers, while preserving the $\mathcal{O}(L)$ computational efficiency of SSMs during inference. This mathematical framework provides the requisite scalability to model long-term procedural contexts.

\begin{figure*}[!t]
\centerline{\includegraphics[width=\textwidth]{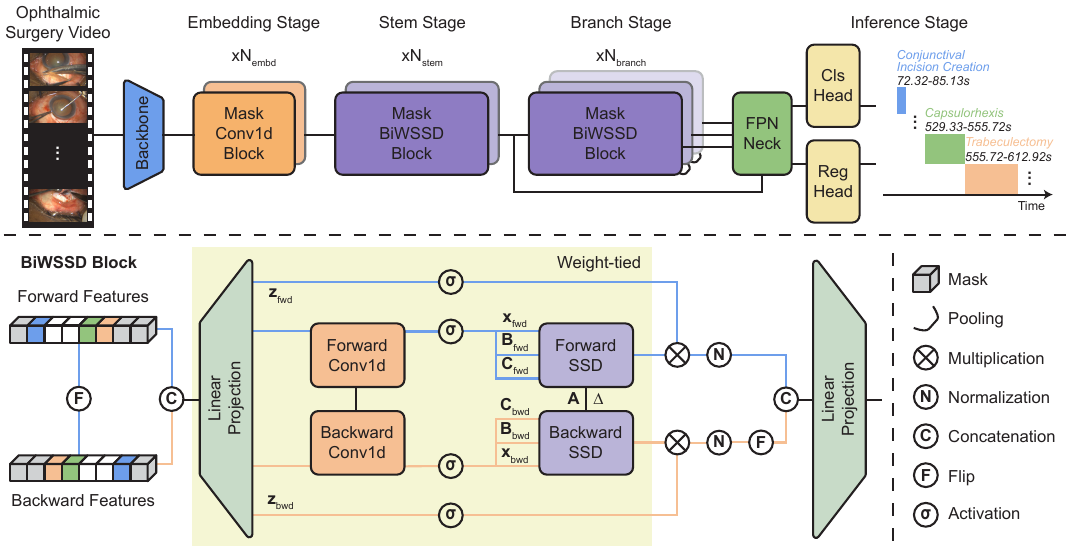}}
\caption{Overview of the OphBiWSSD architecture. The framework processes video features through a hierarchical structure of MaskConv1d and MaskBiWSSD blocks, followed by a unified FPN neck and detection heads for prediction. Each MaskBiWSSD block handles forward and flipped backward features as input. The yellow-shaded region highlights the weight-tied state space duality. $\mathbf{z}$ represents the gating branch to modulate information flow, and $\mathbf{x}$ serves as the input to the state space evolution. The parameters $\mathbf{A}$, $\mathbf{B}$, and $\mathbf{C}$ define the state space latent dynamics, and $\Delta$ controls the discretization step size for temporal scaling.}
\end{figure*}

\begin{algorithm}
\caption{Bidirectional Weight-tied SSD (BiWSSD)}
\label{alg: biwssd}
\begin{algorithmic}[1]
\Require $\mathcal{X} : (B, L, D)$
\Ensure $\mathcal{Y} : (B, L, D)$
\State $\mathbf{A} : (H) \leftarrow \text{Parameter}$

\Comment{Scalar-identity structure per head}
\State $\mathcal{X}_{fwd} \leftarrow \mathcal{X}, \quad \mathcal{X}_{bwd\_rev} \leftarrow \text{Flip}(\mathcal{X})$

\Comment{Bidirectional temporality}
\State $[\mathbf{z}, \mathbf{x}, \mathbf{B}, \mathbf{C}, \Delta] \leftarrow \text{Linear}(\text{Concat}(\mathcal{X}_{fwd}, \mathcal{X}_{bwd\_rev}); \theta_{tied})$

\Comment{Weight-tied projection}
\State $\mathbf{\bar{A}}, \mathbf{\bar{B}} \leftarrow \text{discretize}(\Delta, \mathbf{A}, \mathbf{B})$

\Comment{SSD Discretization}
\State $\mathcal{Y}_{combined} \leftarrow \text{SSD\_Scan}(\mathbf{\bar{A}, \bar{B}, C, x})$

\Comment{Semiseparable matrix multiplication}
\State $\mathcal{Y}_{fwd}, \mathcal{Y}_{bwd\_rev} \leftarrow \text{Split}(\mathcal{Y}_{combined})$

\Comment{Decomposition}
\State $\mathcal{Y}_{bwd} \leftarrow \text{Flip}(\mathcal{Y}_{bwd\_rev})$

\Comment{Temporal restoration}
\State $\mathcal{Y}_{fused} \leftarrow \text{Fuse}(\mathcal{Y}_{fwd}, \mathcal{Y}_{bwd})$

\Comment{Bi-contextual synthesis}
\State $\mathcal{Y} \leftarrow \text{Linear}(\mathcal{Y}_{fused} \circ \sigma(\mathbf{z}))$

\Comment{Non-causal gated projection}

\State \Return $\mathcal{Y}$
\end{algorithmic}
\end{algorithm}

\subsection{Overall Architecture}

The proposed OphBiWSSD architecture maps a high-dimensional sequence of temporal features to a set of discrete action segments with precise temporal boundaries. Formally, given an input feature sequence $\mathcal{X}$ extracted from an ophthalmic surgical video, the model computes the transformation:
\begin{equation}
\mathcal{Y} = \mathcal{H}(\mathcal{X}, \Omega), \quad \mathcal{X} \in \mathbb{R}^{L \times D}, \Omega \in \{0, 1\}^L
\end{equation}

where $L$ denotes the sequence length, $D$ is the feature dimension, and $\Omega$ represents the binary temporal mask utilized to handle variable-length sequences. The operator $\mathcal{H}$ represents the complete hierarchical BiWSSD transformation.

The architecture comprises a multi-stage backbone that transitions from local feature extraction to global procedural modeling. The following sequence of transformations defines this hierarchical extraction of multi-scale temporal features:

\subsubsection{Local Feature Embedding}

The input sequence $\mathcal{X}$ is first projected into an embedding space $\mathbb{R}^{d_{embd}}$ through a series of masked convolutional layers:
\begin{equation}
\mathcal{F}_{0} = \Psi_{embd}(\mathcal{X}, \Omega_{0})
\end{equation}

where $\Psi_{embd}$ consists of a stack of $N_{embd}$ MaskConv1D blocks designed to preserve the initial temporal resolution while encoding local instrument-tissue interactions.

\subsubsection{Constant-Resolution Stem}

To capture global procedural dependencies at the native temporal scale, the embedded features are processed by a series of MaskBiWSSD blocks:
\begin{equation}
\mathcal{F}_{i} = \text{MaskBiWSSD}_{i}(\mathcal{F}_{i-1}, \Omega_{i-1}), \quad i = 1, \dots, N_{stem}
\end{equation}

Each block integrates a BiWSSD (Bidirectional Weight-tied State Space Duality) layer. This layer serves as the core architectural innovation, facilitating non-causal temporal modeling by integrating forward and backward scanning paths. This bidirectional synthesis is critical for resolving the inherent asymmetry in surgical gestures, in which post-action context is as vital as the preceding signal for boundary localization.

\subsubsection{Multi-Scale Branch}

To accommodate actions of varying durations, the architecture generates a feature pyramid through strided downsampling:
\begin{equation}
\begin{aligned}
\mathcal{F}_{j} = \text{MaskBiWSSD}_{j}(\text{DS}(\mathcal{F}_{j-1}, \Omega_{j-1})), \\
\quad j = N_{stem}+1, \dots, N_{branch}
\end{aligned}
\end{equation}

This stage produces a hierarchy of feature maps from each branch, $\{\mathcal{F}_{N_{stem}}, \dots, \mathcal{F}_{N_{branch}}\}$, with progressively reduced temporal resolution.

The efficiency of this backbone is driven by the Weight-tied Selective Scan, in which the selection dynamics of the state-space matrices are governed by a unified weight set, ensuring direction-invariant feature learning. The detailed mathematical derivations for the Bidirectional SSD and the weight-tying mechanism are provided in the subsequent subsections. Finally, the resulting feature hierarchy is passed to a detection head for simultaneous classification and boundary refinement.

The training objective is defined by a multi-task loss function $\mathcal{L}$ that jointly optimizes action classification and temporal boundary regression. This composite loss is formulated as:
\begin{equation}
\begin{aligned}
\mathcal{L} &= \frac{1}{N_{pos}} \sum_{l,t} \mathds{1}_{\{c^l_t>0\}} (\sigma_{IoU} \mathcal{L}_{cls} + \mathcal{L}_{reg}) \\
&+ \frac{1}{N_{neg}} \sum_{l,t} \mathds{1}_{\{c^l_t=0\}} \mathcal{L}_{cls}
\end{aligned}
\end{equation}

Where $\mathds{1}_{\{\cdot\}}$ is the indicator function, the condition $c^l_t > 0$ identifies foreground samples at feature level $l$ and temporal position $t$, where the index $c$ corresponds to a specific surgical action or phase category, $c^l_t = 0$ denotes background samples where no surgical action is present, $\sigma_{IoU}$ denotes the overlap between the predicted segment and ground truth, $\mathcal{L}_{cls}$ and $\mathcal{L}_{reg}$ represent Focal Loss~\cite{lin2017focal} and IoU Loss~\cite{rezatofighi2019generalized}, respectively.

\subsection{Bidirectional Selective Scan Mechanism}

Standard SSMs use recursive formulations to maintain linear complexity, but their inherent causality limits the integration of anti-causal context. To realize a global receptive field without the quadratic scaling of self-attention, we implement a bidirectional selective scan mechanism that aggregates preceding and succeeding surgical cues via bilateral latent state evolution. This architecture governs state transitions via the following discrete-time recurrence relations:
\begin{equation}
\overrightarrow{h}_t = \mathbf{\bar{A}}\overrightarrow{h}_{t-1} + \mathbf{\bar{B}}x_t, \quad \overleftarrow{h}_t = \mathbf{\bar{A}}\overleftarrow{h}_{t+1} + \mathbf{\bar{B}}x_t
\end{equation}

where $\overrightarrow{h}_t$ and $\overleftarrow{h}_t$ represent the forward and backward latent states, respectively. These states are mapped to the output space through the discretized projection matrices $\mathbf{\bar{A}}$, $\mathbf{\bar{B}}$, and $\mathbf{C}$. For computational efficiency, the implementation leverages a hardware-aware optimization by concatenating the original and temporally reversed sequences along the batch dimension, allowing a single kernel call to process both directions. The resulting transformation is defined as:
\begin{equation}
\mathcal{Y}_{fwd} = \text{Scan}(\mathbf{x}, \theta), \quad \mathcal{Y}_{bwd} = \text{Flip}(\text{Scan}(\text{Flip}(\mathbf{x}), \theta))
\end{equation}

where $\text{Flip}(\cdot)$ denotes the temporal reversal operator and $\theta$ represents the state-space parameters. To integrate these multi-directional temporal contexts, the outputs are fused using a parameterizable operator $\mathcal{T}_{\text{fuse}} \in \{\text{concat}, \text{add}\}$ followed by a linear projection:
\begin{equation}
\mathcal{F}_{out} =  \begin{cases}  \mathbf{W}_{out}\left[ \mathcal{Y}_{fwd} \mathbin{\|} \mathcal{Y}_{bwd} \right], & \text{if } \mathcal{T}_{\text{fuse}} = \text{concat} \\ \mathbf{W}_{out}\left(\mathcal{Y}_{fwd} + \mathcal{Y}_{bwd}\right), & \text{if } \mathcal{T}_{\text{fuse}} = \text{add} \end{cases}
\end{equation}

In the concatenation configuration, the learnable projection matrix $\mathbf{W}_{out} \in \mathbb{R}^{2d_{inner} \times d_{embd}}$ compresses the doubled feature dimension back to the model space. Conversely, the summation macro utilizes $\mathbf{W}_{out} \in \mathbb{R}^{d_{inner} \times d_{embd}}$. These fusion strategies enable MaskBiWSSD blocks to synthesize a high-fidelity representation that captures the global bilateral dependencies inherent in dense surgical workflows.

\subsection{Weight-tied State Space Duality}

Surgical maneuvers, such as suturing or incision, exhibit fundamental temporal characteristics that should remain identifiable irrespective of the scanning direction. To enforce this principle within the temporal modeling framework, we introduce a parameter-sharing constraint through Weight-tied State Space Duality. This strategy serves as a structural regularizer, ensuring that the model learns direction-invariant features while mitigating the risk of overfitting on specialized surgical datasets.

In this weight-tied formulation, the motion dynamics for both the forward and backward paths are governed by a unified projection logic. Rather than employing independent linear operators for each temporal direction, a single shared linear transformation $\Phi(\cdot; \theta_{tied})$ is utilized to generate the requisite state-space parameters from the input feature sequence $\mathcal{X}$. The projection is defined as:
\begin{equation}
[\mathbf{z, x, B, C, \mathrm{\Delta}}]_{fwd, bwd} = \Phi(\mathcal{X})
\end{equation}

where $\mathbf{z} \in \mathbb{R}^{d_{inner}}$ serves as the gating branch to modulate information flow, and $\mathbf{x} \in \mathbb{R}^{d_{inner}}$ represents the projected input sequence entering the selective scan. The matrices $\mathbf{B}$ and $\mathbf{C}$ define the state-to-latent and latent-to-output mappings, respectively, while $\Delta$ signifies the learned discretization step size.

By deriving these parameters from a single shared weight set, the model ensures that the $\textit{importance}$ and $\textit{retention}$ of a specific surgical frame are calculated consistently across both temporal contexts. Weight-tied motion dynamics reduce the parameter footprint during the input projection phase while enforcing symmetry in feature extraction. Consequently, the resulting latent representations are more robust to temporal variations, providing a stable foundation for the detection head to perform precise boundary regression and action classification in ophthalmic workflows.

\begin{table}
\label{table_example}
\renewcommand{\arraystretch}{1.3}
\centering
\caption{Architectural and Optimization Hyperparameters}
\label{tab: model settings}
\begin{tabular}{l | l | l}
\hline
\textbf{Category} & \textbf{Parameter} & \textbf{Value} \\
\hline
\textbf{Backbone} & \makecell[l]{Architecture Depth \\ $[N_{embd}, N_{stem}, N_{branch}]$} & $[2, 2, 5]$ \\
& Input Dimension ($D$) & $1408$ \\
& Embedding Dimension ($d_{embd}$) & $512$ \\
& Intermediate Dimension ($d_{inner}$) & $1024$ \\
& Max Sequence Length ($L$) & $2304$ \\
\hline
\textbf{SSD Config} & Number of SSD Heads ($P$) & $32$ \\
\hline
\textbf{Optimization} & Learning Rate ($\eta$) & $1.0 \times 10^{-4}$ \\
& Weight Decay ($\lambda$) & $0.05$ \\
& Warmup Epochs & $20$ \\
& Total Epochs & $40$ \\
\hline
\textbf{Regularization} & DropPath Rate & $0.3$ \\
& Input Noise Variance & $5.0 \times 10^{-4}$ \\
\hline
\end{tabular}
\end{table}

\section{Experiment Setup}

\subsection{Dataset and Task Formulation}

OphNet~\cite{hu2024ophnet} is a large-scale, publicly available benchmark for dense action localization in ophthalmic surgery, intended to advance the field from reliance on small, private datasets to standardized evaluation protocols. The dataset incorporates a hierarchical taxonomy that enables a dual-task approach: predicting action categories $c \in \mathcal{C}$ and identifying precise temporal boundaries $(s, e)$ within continuous surgical sequences. The frequent intraocular tool interactions and the constrained microscopic operative field create substantial ambiguity in temporal transitions, as gesture identity is often determined only by subsequent context. This challenge necessitates non-causal modeling to integrate future surgical cues for accurate boundary refinement.

\subsubsection{Temporal Phase Localization}

This task corresponds to the coarse-grained level of the OphNet hierarchy and requires classification and temporal regression of 52 procedural milestones. It comprises 14,674 instances that require long-range procedural consistency to segment the global surgical workflow accurately. The objective is to identify the primary stages of a procedure, such as ``Phacoemulsification'' or ``IOL Insertion'', which typically span longer temporal durations.

\subsubsection{Temporal Operation Localization}

This fine-grained task focuses on localizing 107 distinct instrument-tissue gestures. With 17,508 instances, the sub-task requires high local precision to differentiate subtle tool interactions from the high-frequency maneuvers characteristic of the ophthalmic domain. The objective is to detect transient technical actions, such as specific suturing maneuvers or capsulorhexis segments, which require superior resolution of temporal boundaries.

\subsection{Evaluation Metrics}

Temporal localization performance on the OphNet benchmark is quantified using mean Average Precision (mAP), which measures both classification accuracy and temporal boundary regression precision. A predicted segment $[s_p, e_p]$ is considered as a true positive when its action category aligns with the ground truth and the temporal Intersection over Union ($tIoU$) between the prediction and the ground truth segment $[s_g, e_g]$ exceeds a predefined threshold $\alpha$. The $tIoU$ is formally defined as the ratio of the intersection to the union of the two temporal intervals:

\begin{equation}
tIoU = \frac{|[s_p, e_p] \cap [s_g, e_g]|}{|[s_p, e_p] \cup [s_g, e_g]|}
\end{equation}

The Average Precision (AP) for each surgical gesture category is calculated by ranking predictions by their confidence scores to produce a Precision-Recall curve. Higher $tIoU$ thresholds impose stringent requirements on the model to capture fine-grained entry and exit boundaries of surgical instruments, a capability essential for clinical skill assessment and intraoperative safety monitoring. To provide a comprehensive assessment across all surgical action categories, the mAP is computed as the arithmetic mean of the $AP$ values across the total number of categories $C$:

\begin{equation}
mAP = \frac{1}{C} \sum_{c=1}^{C} AP_c
\end{equation}

Following the established benchmarking standards for OphNet~\cite{hu2024ophnet}, mAP is reported at multiple $tIoU$ thresholds, typically $\alpha \in \{0.3, 0.4, 0.5, 0.6, 0.7\}$, to characterize model robustness across varying degrees of boundary precision. The average mAP computed across the predefined thresholds constitutes the primary performance metric for evaluating the overall efficacy of the model in TAL tasks.

\begin{table*}[!t]
\renewcommand{\arraystretch}{1.3}
\caption{Performance comparison For temporal phase localization and temporal operation localization. The model with the symbol $\dag$ represents using the same detection head and loss function as in~\cite{li2025temporal}.}
\resizebox{\textwidth}{!}{%
\begin{tabular}{c|c|cccccc|cccccc}
\hline
 &  & \multicolumn{6}{c|}{Temporal Phase Localization mAP (\%)} & \multicolumn{6}{c}{Temporal Operation Localization mAP (\%)} \\ \cline{3-14} 
\multirow{-2}{*}{Model} & \multirow{-2}{*}{Backbone} & 0.3 & 0.4 & 0.5 & 0.6 & 0.7 & Avg. & 0.3 & 0.4 & 0.5 & 0.6 & 0.7 & Avg. \\ \hline
 & CSN~\cite{tran2019video} & 31.62 & 28.89 & 24.88 & 20.56 & 14.99 & 24.19 & 29.75 & 26.95 & 23.21 & 17.77 & 13.42 & 22.22 \\
 & SlowFast~\cite{feichtenhofer2019slowfast} & 33.06 & 30.44 & 26.11 & 20.51 & 14.25 & 24.88 & 30.75 & 28.66 & 24.91 & 18.93 & 14.26 & 23.50 \\
 & SwinViviT~\cite{liu2022video} & 37.28 & 34.32 & 29.61 & 24.08 & 18.67 & 28.79 & 33.87 & 30.11 & 26.14 & 20.26 & 13.97 & 24.87 \\
\multirow{-4}{*}{\begin{tabular}[c]{@{}c@{}}\color{textgray}{[}ECCV 2022{]}\\ ActionFormer~\cite{zhang2022actionformer}\end{tabular}} & \cellcolor{tablegray}VideoMAE~\cite{wang2023videomae} & \cellcolor{tablegray}46.49 & \cellcolor{tablegray}43.61 & \cellcolor{tablegray}38.62 & \cellcolor{tablegray}33.33 & \cellcolor{tablegray}26.04 & \cellcolor{tablegray}37.62 & \cellcolor{tablegray}45.44 & \cellcolor{tablegray}42.68 & \cellcolor{tablegray}37.92 & \cellcolor{tablegray}31.46 & \cellcolor{tablegray}24.61 & \cellcolor{tablegray}36.42 \\ \hline
 & CSN & 32.58 & 29.92 & 25.82 & 21.50 & 14.13 & 24.79 & 33.64 & 30.30 & 27.33 & 22.47 & 16.73 & 26.09 \\
 & SlowFast & 34.50 & 32.26 & 28.40 & 23.25 & 17.60 & 27.20 & 35.54 & 32.87 & 28.88 & 24.10 & 17.47 & 27.77 \\
 & SwinViviT & 38.62 & 35.13 & 30.85 & 25.16 & 20.51 & 30.06 & 35.00 & 32.03 & 27.86 & 23.37 & 16.86 & 27.02 \\
\multirow{-4}{*}{\begin{tabular}[c]{@{}c@{}}\color{textgray}{[}CVPR 2023{]}\\ TriDet~\cite{shi2023tridet}\end{tabular}} & \cellcolor{tablegray}VideoMAE & \cellcolor{tablegray}46.70 & \cellcolor{tablegray}44.23 & \cellcolor{tablegray}41.33 & \cellcolor{tablegray}35.22 & \cellcolor{tablegray}28.61 & \cellcolor{tablegray}39.22 & \cellcolor{tablegray}48.45 & \cellcolor{tablegray}46.02 & \cellcolor{tablegray}41.36 & \cellcolor{tablegray}35.82 & \cellcolor{tablegray}29.69 & \cellcolor{tablegray}40.27 \\ \hline
 & CSN & 34.14 & 30.50 & 27.08 & 21.53 & 16.16 & 25.88 & 30.92 & 27.70 & 23.79 & 19.14 & 14.15 & 23.14 \\
 & SlowFast & 34.99 & 31.86 & 28.59 & 21.52 & 16.80 & 26.75 & 30.37 & 27.35 & 24.42 & 19.60 & 15.28 & 23.41 \\
 & SwinViviT & 38.52 & 35.51 & 31.37 & 26.17 & 19.76 & 30.27 & 35.04 & 31.65 & 28.51 & 23.19 & 17.84 & 27.25 \\
\multirow{-4}{*}{\begin{tabular}[c]{@{}c@{}}\color{textgray}{[}ECCV 2024{]}\\ DyFADet~\cite{yang2024dyfadet}\end{tabular}} & \cellcolor{tablegray}VideoMAE & \cellcolor{tablegray}47.82 & \cellcolor{tablegray}45.30 & \cellcolor{tablegray}39.74 & \cellcolor{tablegray}34.24 & \cellcolor{tablegray}28.58 & \cellcolor{tablegray}39.14 & \cellcolor{tablegray}46.20 & \cellcolor{tablegray}43.98 & \cellcolor{tablegray}40.12 & \cellcolor{tablegray}33.86 & \cellcolor{tablegray}27.48 & \cellcolor{tablegray}38.33 \\ \hline
 & CSN & 35.48 & 32.34 & 28.50 & 23.51 & 18.63 & 27.69 & 32.24 & 29.23 & 26.76 & 22.50 & 16.51 & 25.45 \\
 & SlowFast & 36.93 & 33.95 & 29.41 & 24.86 & 19.22 & 28.87 & 34.13 & 31.71 & 28.01 & 23.35 & 17.99 & 27.04 \\
 & SwinViviT & 37.61 & 35.06 & 30.73 & 25.39 & 19.57 & 29.67 & 36.47 & 33.80 & 29.84 & 25.69 & 19.37 & 29.04 \\
\multirow{-4}{*}{\begin{tabular}[c]{@{}c@{}}\color{textgray}{[}AAAI 2025{]}\\ CLTDR-GMG~\cite{li2025temporal}\end{tabular}} & \cellcolor{tablegray}VideoMAE & \cellcolor{tablegray}50.24 & \cellcolor{tablegray}47.39 & \cellcolor{tablegray}43.18 & \cellcolor{tablegray}37.82 & \cellcolor{tablegray}31.41 & \cellcolor{tablegray}42.01 & \cellcolor{tablegray}48.95 & \cellcolor{tablegray}45.51 & \cellcolor{tablegray}41.19 & \cellcolor{tablegray}36.67 & \cellcolor{tablegray}29.68 & \cellcolor{tablegray}40.40 \\ \hline
 & CSN & 36.14 & 33.60 & 29.97 & 22.77 & 16.25 & 27.75 & 35.38 & 31.83 & 27.74 & 22.20 & 17.03 & 26.84 \\
 & SlowFast & 39.05 & 36.00 & 31.30 & 25.13 & 19.24 & 30.14 & 39.21 & 36.06 & 31.29 & 25.36 & 19.85 & 30.36 \\
 & SwinViviT & 41.03 & 36.82 & 31.48 & 24.02 & 19.21 & 30.51 & 37.05 & 33.84 & 29.02 & 22.69 & 15.94 & 27.71 \\
\multirow{-4}{*}{\begin{tabular}[c]{@{}c@{}}\color{textgray}{[}IJCV 2026{]}\\ ActionMamba~\cite{chen2026video}\end{tabular}} & \cellcolor{tablegray}VideoMAE & \cellcolor{tablegray}49.53 & \cellcolor{tablegray}47.37 & \cellcolor{tablegray}42.65 & \cellcolor{tablegray}35.77 & \cellcolor{tablegray}27.32 & \cellcolor{tablegray}40.53 & \cellcolor{tablegray}49.18 & \cellcolor{tablegray}46.34 & \cellcolor{tablegray}42.05 & \cellcolor{tablegray}34.71 & \cellcolor{tablegray}26.87 & \cellcolor{tablegray}39.83 \\ \hline
 & CSN & 38.21 & 35.25 & 31.62 & 24.99 & 19.69 & 29.95 & 39.08 & 34.83 & 31.04 & 25.56 & 20.33 & 30.17 \\
 & SlowFast & 40.99 & 36.86 & 32.19 & 26.67 & 21.42 & 31.63 & 41.66 & 38.55 & 34.01 & 28.60 & 21.85 & 32.93 \\
 & SwinViviT & 41.13 & 38.16 & 33.19 & 29.31 & 22.25 & 32.81 & 41.59 & 38.69 & 33.71 & 27.76 & 22.42 & 32.83 \\
\multirow{-4}{*}{\textbf{OphBiWSSD (Ours)}} & \cellcolor{tablegray}VideoMAE & \cellcolor{tablegray}\textbf{52.98} & \cellcolor{tablegray}\textbf{50.19} & \cellcolor{tablegray}\textbf{45.76} & \cellcolor{tablegray}\textbf{39.97} & \cellcolor{tablegray}\textbf{33.18} & \cellcolor{tablegray}\textbf{44.42} & \cellcolor{tablegray}\textbf{52.59} & \cellcolor{tablegray}\textbf{49.22} & \cellcolor{tablegray}\textbf{45.13} & \cellcolor{tablegray}\textbf{37.83} & \cellcolor{tablegray}\textbf{30.62} & \cellcolor{tablegray}\textbf{43.08} \\ \hline
 & CSN & \multicolumn{1}{l}{37.87} & \multicolumn{1}{l}{35.14} & \multicolumn{1}{l}{30.28} & \multicolumn{1}{l}{25.39} & \multicolumn{1}{l}{20.61} & \multicolumn{1}{l|}{29.86} & \multicolumn{1}{l}{38.94} & \multicolumn{1}{l}{35.86} & \multicolumn{1}{l}{31.75} & \multicolumn{1}{l}{26.81} & \multicolumn{1}{l}{21.56} & \multicolumn{1}{l}{30.98} \\
 & SlowFast & \multicolumn{1}{l}{41.74} & \multicolumn{1}{l}{39.07} & \multicolumn{1}{l}{34.55} & \multicolumn{1}{l}{29.28} & \multicolumn{1}{l}{23.20} & \multicolumn{1}{l|}{33.57} & \multicolumn{1}{l}{41.85} & \multicolumn{1}{l}{38.49} & \multicolumn{1}{l}{34.54} & \multicolumn{1}{l}{29.01} & \multicolumn{1}{l}{23.00} & \multicolumn{1}{l}{33.38} \\
 & SwinViviT & \multicolumn{1}{l}{42.31} & \multicolumn{1}{l}{39.12} & \multicolumn{1}{l}{35.15} & \multicolumn{1}{l}{29.33} & \multicolumn{1}{l}{23.22} & \multicolumn{1}{l|}{33.83} & \multicolumn{1}{l}{39.05} & \multicolumn{1}{l}{36.55} & \multicolumn{1}{l}{31.95} & \multicolumn{1}{l}{27.26} & \multicolumn{1}{l}{21.23} & \multicolumn{1}{l}{31.21} \\
\multirow{-4}{*}{\textbf{OphBiWSSD\dag (Ours)}} & \cellcolor{tablegray}VideoMAE & \multicolumn{1}{l}{\cellcolor{tablegray}\textbf{54.76}} & \multicolumn{1}{l}{\cellcolor{tablegray}\textbf{51.55}} & \multicolumn{1}{l}{\cellcolor{tablegray}\textbf{46.41}} & \multicolumn{1}{l}{\cellcolor{tablegray}\textbf{41.32}} & \multicolumn{1}{l}{\cellcolor{tablegray}\textbf{34.96}} & \multicolumn{1}{l|}{\cellcolor{tablegray}\textbf{45.80}} & \multicolumn{1}{l}{\cellcolor{tablegray}\textbf{52.60}} & \multicolumn{1}{l}{\cellcolor{tablegray}\textbf{49.83}} & \multicolumn{1}{l}{\cellcolor{tablegray}\textbf{46.35}} & \multicolumn{1}{l}{\cellcolor{tablegray}\textbf{39.80}} & \multicolumn{1}{l}{\cellcolor{tablegray}\textbf{33.46}} & \multicolumn{1}{l}{\cellcolor{tablegray}\textbf{44.41}} \\ \hline
\end{tabular}%
}
\end{table*}

\subsection{Implementation Details}

The proposed OphBiWSSD framework is implemented in PyTorch and optimized for high-dimensional surgical video analysis. All experiments are conducted in a single NVIDIA GeForce RTX 4090 GPU. The input feature sequence $\mathcal{X}$ is extracted from ophthalmic surgical videos using diverse backbones, including CSN, SlowFast, SwinViviT, and VideoMAEv2-g~\cite{wang2023videomae}, to evaluate the performance of the model across varying spatio-temporal representations. During the training phase, a maximum sequence length of $L = 2304$ is utilized to accommodate long procedural dependencies characteristic of complex ophthalmic interventions.

The model uses a hierarchical backbone with an architectural depth of $[2, 2, 5]$, corresponding to the number of layers in the Local Feature Embedding, Constant-Resolution Stem, and Multi-Scale Branch stages, respectively. Each MaskBiWSSD block is configured with an embedding dimension $d_{embd} = 512$. For the SSD mechanism, we set the number of SSD heads $P = 32$. To enhance model robustness and prevent overfitting on the specialized OphNet dataset, we set the stochastic depth (DropPath~\cite{huang2016deep}) rate to $0.3$ and inject Gaussian input noise with a variance of $5.0 \times 10^{-4}$.

The localization component adopts a Trident-head architecture~\cite{shi2023tridet} that performs simultaneous classification and boundary regression across multiple feature scales. This head incorporates a bin-based refinement strategy with $num\_bins = 16$ to achieve sub-grid localization precision. The total loss is the sum of classification and regression losses. With $\lambda_{cls}$ and $\lambda_{reg}$ both set to 1.0, accuracy and boundary precision are equally important in the ophthalmic surgical TAL task. Post-processing is conducted via batched Soft Non-Maximum Suppression (Soft-NMS)~\cite{bodla2017soft} with a voting threshold of $0.7$. The complete set of optimization and architectural parameters is summarized in Table~\ref{tab: model settings}.

\section{Results}

\subsection{Temporal Phase Localization}

OphBiWSSD achieves state-of-the-art performance on OphNet for temporal phase localization, outperforming prior models across all $tIoU$ thresholds. The model maintains strong performance across various feature representations, exhibiting effective synergy with transformer-based feature backbones and outperforming CNN-based counterparts. With the VideoMAEv2-g backbone, OphBiWSSD attains an average mAP of 44.42\%, while the enhanced OphBiWSSD$\dag$ variant reaches 45.80\%. Notably, our model surpasses ActionFormer and TriDet by 6.80\% and 5.20\%, respectively. Its advantage becomes even more pronounced at higher $tIoU$ thresholds, underscoring the model’s precision in delineating surgical phase boundaries. At $\alpha=0.7$, OphBiWSSD achieves an average mAP of 33.18\%, significantly exceeding ActionFormer (26.04\%) and TriDet (28.61\%). Additionally, OphBiWSSD outperforms the Mamba-based ActionMamba by 3.89\%. When using the same detection head and loss function as CLTDR-GMG, OphBiWSSD$\dag$ delivers a 3.79\% improvement, highlighting the robustness of the SSD backbone for modeling long-range dependencies in surgical workflow analysis.

\subsection{Temporal Operation Localization}

The temporal operation localization task on the OphNet benchmark presents greater complexity than phase localization due to the presence of 107 distinct action categories, each defined by high-frequency maneuvers and brief durations. Within this fine-grained and challenging context, OphBiWSSD demonstrates an empirically superior average mAP of 43.08\%, marking a 6.66\% improvement over established benchmarks. The model sustains high performance under stringent evaluation criteria, achieving an mAP of 30.62\% at the $\alpha=0.7$ threshold. Robustness at elevated $tIoU$ levels is essential for accurately identifying the entry and exit points of instruments, which is necessary for objective clinical skill assessment. A comparative analysis of spatio-temporal representations reveals that integrating the VideoMAEv2-g backbone with the BiWSSD framework produces the most robust localization outcomes, effectively accommodating the high-dimensional feature sequences characteristic of dense surgical workflows. Furthermore, the enhanced variant OphBiWSSD$\dag$, which attains an average mAP of 44.41\%, demonstrates a statistically significant 4.01\% improvement over CLTDR-GMG at 40.40\%.

\begin{table}
\renewcommand{\arraystretch}{1.3}
\centering
\caption{Ablation study on the bidirectional strategy.}
\label{tab: bidirectional strategy}
\resizebox{\columnwidth}{!}{%
\begin{tabular}{c|cccccc}
\hline
Bidirectional Strategy & 0.3 & 0.4 & 0.5 & 0.6 & 0.7 & Avg. \\ \hline
Forward Only & 48.48 & 45.19 & 41.20 & 35.57 & 28.93 & 39.87 \\
Bidirectional Sequence & 47.53 & 44.19 & 40.10 & 35.40 & 28.53 & 39.15 \\
Bidirectional Block & 47.72 & 44.55 & 40.27 & 35.58 & 29.83 & 39.59 \\
\textbf{Bidirectional SSD} & \textbf{49.90} & \textbf{46.71} & \textbf{41.95} & \textbf{37.25} & \textbf{30.97} & \textbf{41.36} \\ \hline
\end{tabular}%
}
\end{table}

\begin{table}
\renewcommand{\arraystretch}{1.3}
\centering
\caption{Ablation study on the weight-tied strategy.}
\label{tab: weight-tied strategy}
\resizebox{\columnwidth}{!}{%
\begin{tabular}{c|cc|c}
\hline
Weight-tied Strategy & Params (M) & Latency (ms) & Avg. mAP (\%) \\ \hline
Fully Untied & 3.456 & 2.266 & 41.33 \\
Tied Proj+Untied SSD & 2.260 & 2.315 & 41.48 \\
Untied Proj+Tied SSD & 3.448 & 1.255 & \textbf{41.93} \\
\textbf{Weight-Tied SSD} & \textbf{2.252} & \textbf{1.251} & 41.36 \\ \hline
\end{tabular}%
}
\end{table}

\begin{figure*}[!t]
\centerline{\includegraphics[width=\textwidth]{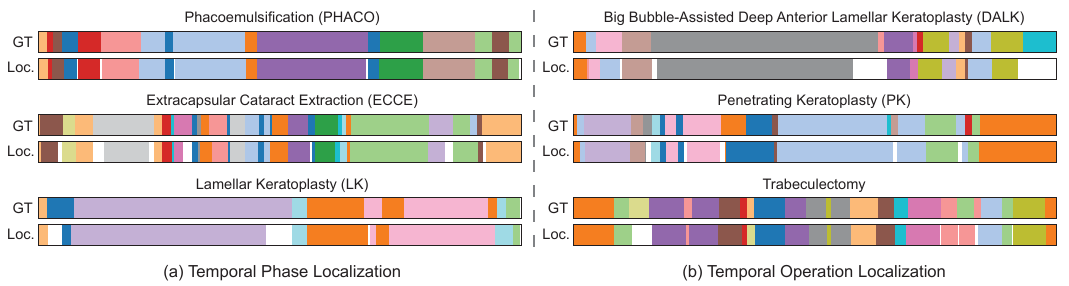}}
\caption{Qualitative results of OphBiWSSD on temporal phase localization and temporal operation localization. GT displays the ground-truth visualization of actions with their accurate temporal boundaries, Loc. shows the most confident actions with their predicted temporal boundaries. Colored segments represent distinct action categories. Blank segments indicate invalid regions or gaps in action.}
\label{fig: qualitative results}
\end{figure*}

\subsection{Ablation Study}

\subsubsection{Bidirectional Strategy}

The necessity of non-causal temporal synthesis is highlighted by the comparative analysis of scanning directions presented in Table~\ref{tab: bidirectional strategy}. Inspired by Vision Mamba~\cite{zhu2024vision}, the Forward Only baseline is evaluated as a standard causal configuration that processes information strictly in the forward temporal direction. The Bidirectional Sequence strategy applies temporal data augmentation by randomly flipping input sequence pairs during training. In the Bidirectional Block configuration, the architecture employs paired stacked model blocks: the first block in each pair processes the sequence forward, while the second processes it in reverse. The proposed Bidirectional SSD achieves a 1.49\% absolute improvement in average mAP over the causal baseline, underscoring the importance of incorporating anti-causal context within latent state evolution. The observed performance degradation in the Bidirectional Sequence and Bidirectional Block variants indicates that stochastic temporal flipping introduces excessive noise into the high-frequency surgical signal, and block-level reversal does not sufficiently fuse bilateral cues at a fine enough scale to resolve precise gesture boundaries.

\subsubsection{Weight-tied Strategy}

The evaluation of weight-tying strategies, as summarized in Table~\ref{tab: weight-tied strategy}, demonstrates a critical trade-off between representational capacity and computational tractability. The Weight-Tied SSD configuration achieves notable efficiency gains, reducing the parameter footprint by approximately 34.8\% and inference latency by 44.8\% compared to the fully untied baseline. The proposed architecture significantly reduces model complexity while maintaining performance parity with the untied variant. Although the Untied Proj+Tied SSD configuration achieves the highest localization accuracy, with a 0.57\% absolute mAP improvement over the proposed model, it requires a 34.7\% increase in parameter overhead. Therefore, the Weight-Tied SSD represents an optimal solution for intraoperative environments, serving as a structural regularizer that enforces direction-invariant feature learning while maximizing real-time throughput.

\begin{table}
\renewcommand{\arraystretch}{1.3}
\centering
\caption{Ablation study on the bidirectional fusion.}
\label{tab: bidirectional fusion}
\resizebox{\columnwidth}{!}{%
\begin{tabular}{c|cccccc}
\hline
Fusion Strategy & 0.3 & 0.4 & 0.5 & 0.6 & 0.7 & Avg. \\ \hline
Summation & 48.81 & 45.63 & 41.49 & 36.12 & 30.00 & 40.41 \\
\textbf{Concatenation} & \textbf{49.90} & \textbf{46.71} & \textbf{41.95} & \textbf{37.25} & \textbf{30.97} & \textbf{41.36} \\ \hline
\end{tabular}
}
\end{table}

\begin{table}
\renewcommand{\arraystretch}{1.3}
\centering
\caption{Ablation study on the number of SSD heads.}
\label{tab: number of SSD heads}
\begin{tabular}{c|cccccc}
\hline
\# SSD Head & 0.3 & 0.4 & 0.5 & 0.6 & 0.7 & Avg. \\ \hline
16 & 49.45 & 46.52 & \textbf{42.20} & 36.65 & \textbf{31.11} & 41.19 \\
\textbf{32} & \textbf{49.90} & \textbf{46.71} & 41.95 & \textbf{37.25} & 30.97 & \textbf{41.36} \\
64 & 49.31 & 46.67 & 42.07 & 36.90 & 29.44 & 40.88 \\ \hline
\end{tabular}
\end{table}

\subsubsection{Bidirectional Feature Fusion}

The architectural impact of bidirectional feature fusion is evaluated in Table~\ref{tab: bidirectional fusion}. Empirical results demonstrate the superiority of the concatenation strategy, which achieves a 0.95\% absolute improvement in average mAP over the summation baseline. This performance gain suggests that concatenation more effectively preserves the unique semantic signatures derived from the forward and backward scanning paths, whereas element-wise summation likely introduces feature conflation, obscuring the fine-grained transitions critical for ophthalmic TAL.

\subsubsection{Number of SSD Heads}

The model's sensitivity to the granularity of latent states is assessed by varying the number of SSD heads ($P$), as presented in Table~\ref{tab: number of SSD heads}. The results demonstrate that peak performance occurs at $P=32$, yielding a 0.17\% absolute improvement in average mAP compared to the $P=16$ variant and a 0.48\% advantage over the $P=64$ configuration. These findings establish $P=32$ as the optimal set point for capturing the diversity of surgical gestures. In contrast, a lower head count limits the model's representational capacity, while an excessive number of heads leads to over-parameterization and may result in overfitting to the specialized motion dynamics of ophthalmic procedures.

\begin{table}
\renewcommand{\arraystretch}{1.3}
\centering
\caption{Ablation study on the detection head.}
\label{tab: detection head}
\resizebox{\columnwidth}{!}{%
\begin{tabular}{c|cccccc}
\hline
Detection Head & 0.3 & 0.4 & 0.5 & 0.6 & 0.7 & Avg. \\ \hline
Conv Head~\cite{zhang2022actionformer} & 49.90 & 46.71 & 41.95 & 37.25 & 30.97 & 41.36 \\
Trident Head~\cite{shi2023tridet} & 51.63 & 48.77 & 44.56 & 39.91 & 32.43 & 43.46 \\
\textbf{CLTDR}~\cite{li2025temporal} & \textbf{51.66} & \textbf{49.12} & \textbf{45.32} & \textbf{40.58} & \textbf{32.62} & \textbf{43.86} \\ \hline
\end{tabular}%
}
\end{table}

\begin{table}
\renewcommand{\arraystretch}{1.3}
\centering
\caption{Computational analysis of competitive models.}
\label{tab: computational analysis}
\begin{tabular}{c|ccc}
\hline
Model & Params (M) & MACs (G) & Latency (ms) \\ \hline
ActionFormer~\cite{zhang2022actionformer} & 28.26 & 42.99 & 32.36 \\
TriDet~\cite{shi2023tridet} & 15.12 & 42.08 & 26.29 \\
DyFADet~\cite{yang2024dyfadet} & 27.59 & 98.60 & 53.21 \\
CLTDR-GMG~\cite{li2025temporal} & 21.28 & 63.52 & 39.58 \\
ActionMamba~\cite{chen2026video} & 17.62 & 36.95 & 17.66 \\ \hline
OphBiWSSD (Ours) & 21.95 & 57.52 & 23.06 \\ \hline
\end{tabular}
\end{table}

\subsubsection{Detection Head}

The influence of the detection head on temporal localization accuracy is evaluated in Table~\ref{tab: detection head}. The results demonstrate that replacing the baseline convolutional head with the Trident head yields a 2.10\% absolute increase in average mAP, whereas adopting the CLTDR head provides the most significant performance boost, with a 2.50\% lead over the baseline. These gains are primarily attributed to the task decoupling and multi-scale refinement capabilities of the advanced heads, which more effectively meet the high-precision demands of surgical gesture boundary regression.

\subsubsection{Computational Analysis}

The computational efficiency of the model is evaluated in Table~\ref{tab: computational analysis}. The framework establishes a superior accuracy-to-latency balance, achieving a 2.3$\times$ speedup over DyFADet and reducing Multiply-Accumulate operations by 41.7\%. Compared to CLTDR-GMG, OphBiWSSD reduces inference latency by 41.7\% and computational overhead by 9.4\%. Additionally, OphBiWSSD exhibits 12.3\% lower latency than TriDet and a 22.3\% smaller parameter footprint than ActionFormer. Although ActionMamba provides a lighter profile, its capacity to model complex ophthalmic workflows is limited. In contrast, OphBiWSSD achieves absolute improvements of 3.89\% in average mAP for phases and 3.25\% for operations relative to the Mamba baseline.

\subsection{Qualitative Visualization}

\begin{figure*}[!t]
\centerline{\includegraphics[width=0.8\textwidth]{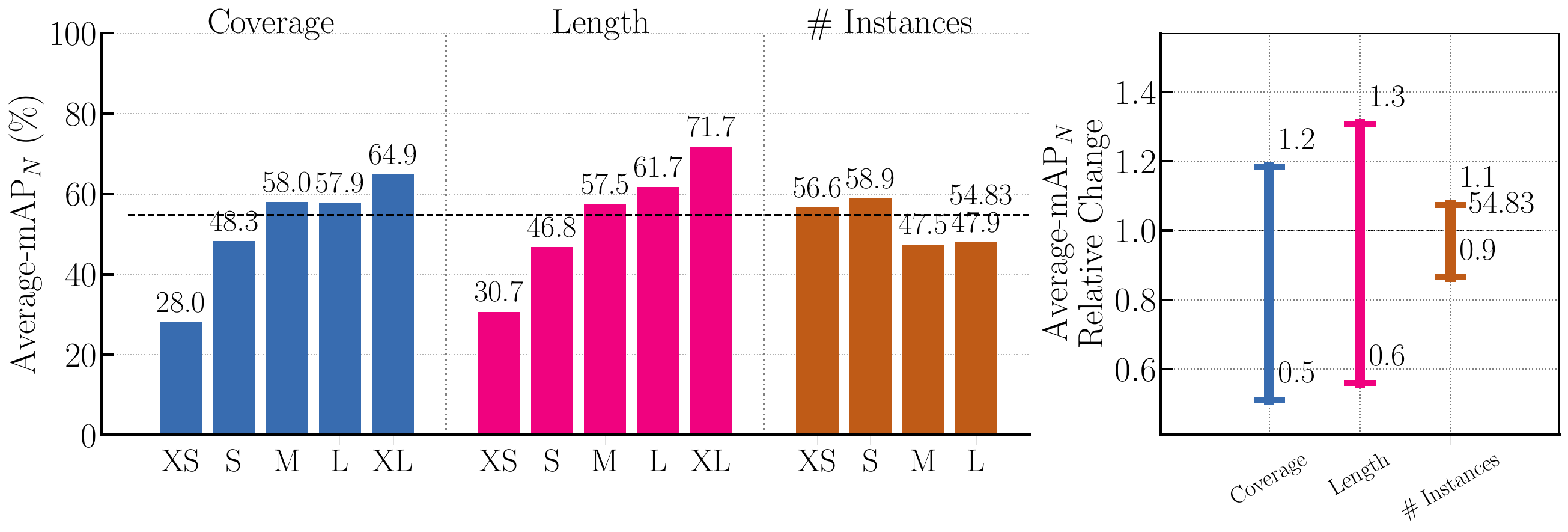}}
\caption{Sensitivity analysis characterizing the performance of OphBiWSSD on OphNet. The average $mAP_N$ is reported for different subsets of ground-truth instances based on their coverage (relative duration), length (absolute duration), and frequency (\# instances) per video.}
\label{fig: sensitivity profile}
\end{figure*}

Qualitative results on OphNet visually demonstrate OphBiWSSD’s effectiveness in capturing the full spectrum of ophthalmic surgical actions. As illustrated in Fig.~\ref{fig: qualitative results}, distinct color-coding highlights precise temporal alignment and strong temporal overlap ($tIoU$). The model consistently maintains accurate category assignments and identification of the onset and offset of surgical maneuvers, even in complex or dense sequences, and handles blank segments robustly. These observations reinforce quantitative findings, confirming the precision and viability of the proposed method in high-fidelity surgical workflow analysis.

\subsection{Error Analysis}

\begin{figure}
\centerline{\includegraphics[width=\columnwidth]{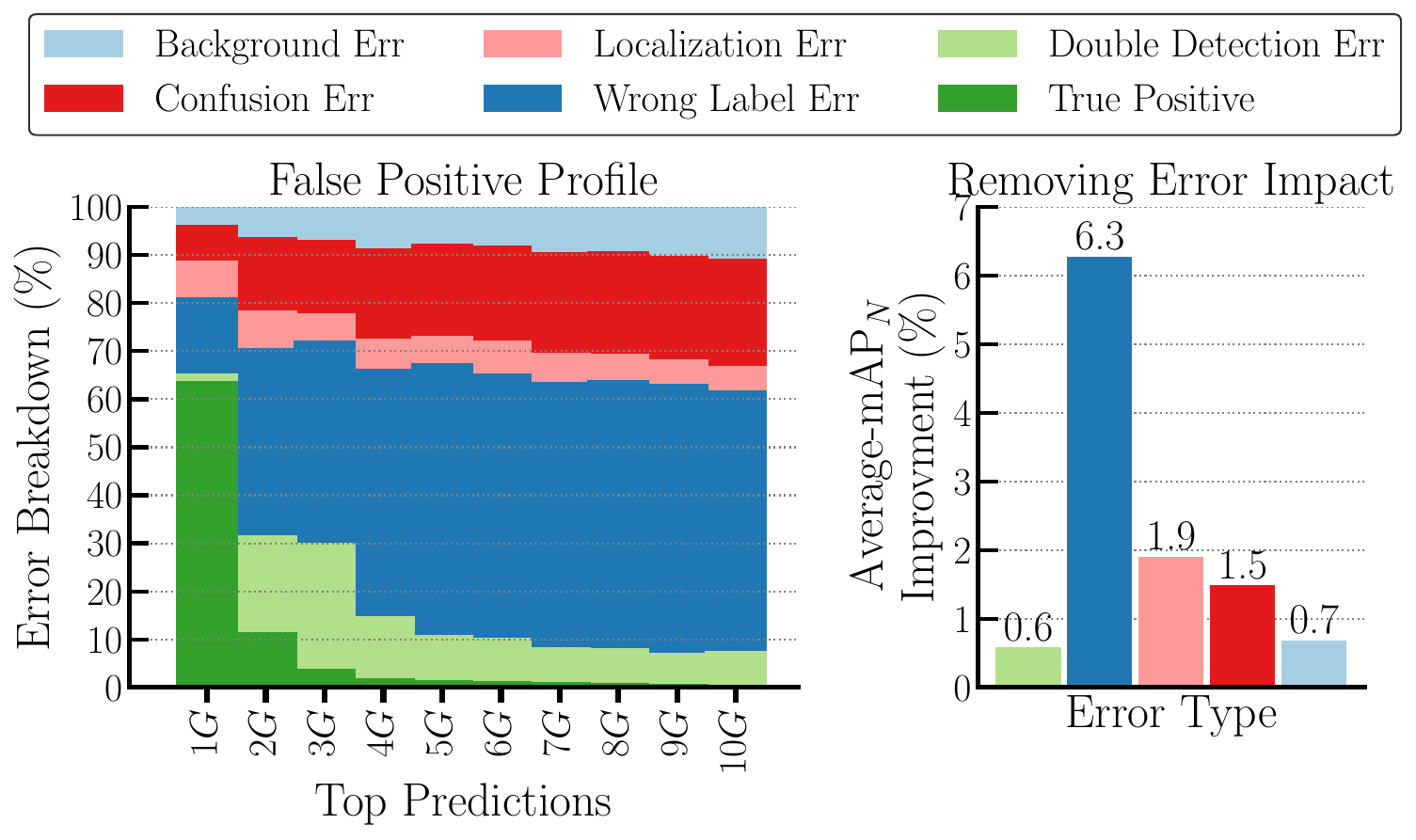}}
\caption{False positive analysis of OphBiWSSD on OphNet. The error breakdown illustrates the distribution of true positives and five distinct error categories across the top predictions. The impact of each error type on the average $mAP_N$ is quantified.}
\label{fig: false positive profile}
\end{figure}

\begin{figure}
\centerline{\includegraphics[width=\columnwidth]{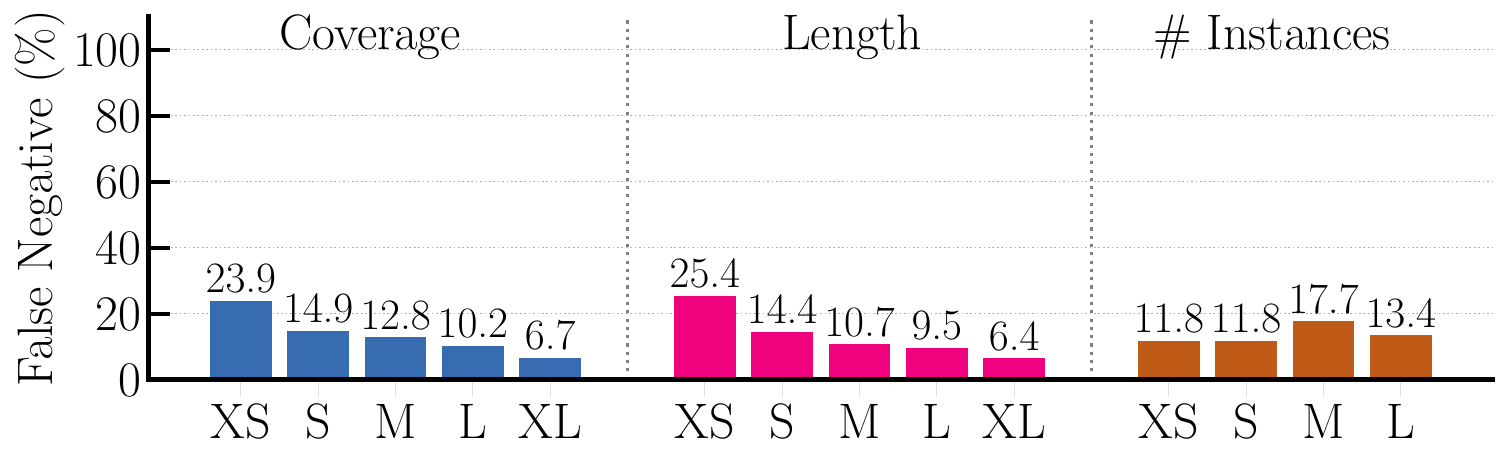}}
\caption{False negative analysis of OphBiWSSD on OphNet depicting the miss detection rate. The percentage of false negatives is categorized by coverage, length, and instance frequency.}
\label{fig: false negative profile}
\end{figure}

To comprehensively evaluate the performance of OphBiWSSD on OphNet, a multi-dimensional error analysis is conducted using the DETAD~\cite{alwassel2018diagnosing} framework. This diagnostic approach extends beyond single-scalar metrics by systematically analyzing failure modes and examining how specific ground-truth instance characteristics influence model robustness. The analysis focuses on normalized mean Average Precision ($mAP_N$), which provides a robust baseline for comparing uneven subsets of ground-truth instances by normalizing precision to a constant number of instances per class. Instances are categorized according to three primary attributes: Coverage, defined as the instance duration relative to the total video length; Length, which denotes the absolute duration in seconds; and \# Instances, indicating the frequency of action occurrences from the same category within a single video.

\subsubsection{Sensitivity Profile}

The sensitivity profile presented in Fig.~\ref{fig: sensitivity profile} quantifies variations in average precision relative to specific ground-truth characteristics, identifying the action attributes that most significantly affect detector performance. Analysis of Coverage and Length indicates that the model is highly effective at localizing long-range surgical actions, but it demonstrates reduced performance for transient gestures with minimal temporal footprints. Furthermore, the model shows increased sensitivity and lower average precision in the M and L categories of instance frequency, indicating that multiple similar gestures or intermediate-duration actions introduce temporal ambiguity that hinders precise boundary localization.

\subsubsection{False Positive Profile}

The false positive profile in Fig.~\ref{fig: false positive profile} facilitates a systematic dissection of error sources by prediction confidence, enabling a thorough evaluation of the detector's classification and scoring components. The primary failure mode of the model is identified as Wrong Label Err, responsible for a notable 6.3\% reduction in performance that could be recovered by eliminating this error. Furthermore, the model exhibits strong localization capability, as evidenced by the minimal impact of Localization Err on performance.

\subsubsection{False Negative Profile}

The false negative profile in Fig.~\ref{fig: false negative profile} provides insights into the types of instances the algorithm struggles to retrieve. Extra Small (XS) instances exhibit high miss rates, at 23.9\% for Coverage and 25.4\% for Length, and drop significantly to 6.7\% and 6.4\% for Extra Large (XL) instances, respectively. This disparity confirms that while the model adeptly captures global procedural contexts, the localization of atomic instrument-tissue interactions remains constrained by the limited visual evidence available in sub-second surgical gesture transitions.

\section{Discussion}

OphBiWSSD achieves state-of-the-art performance on the OphNet benchmark, attaining 44.42\% and 43.08\% mAP for phase and operation localization, respectively. These results significantly surpass established baselines such as ActionFormer and TriDet, and exceed high-performing models including CLTDR-GMG and the Mamba-based ActionMamba. Mechanism validation indicates that the bidirectional selective scan provides a 1.49\% mAP improvement by capturing anti-causal context, while weight-tying enables a 34.8\% reduction in parameters without compromising representational capacity. Qualitative visualizations further confirm that these innovations maintain precise temporal alignment, even during high-frequency maneuvers in complex surgical sequences.

Despite these advancements, critical evaluation identifies Wrong Label Err as the primary performance bottleneck, accounting for a 6.3\% reduction in $mAP_N$. Additionally, the framework exhibits sensitivity to boundary transitions in Extra Small (XS) instances, where miss rates exceed 23\% due to the inherent difficulty of localizing brief gestures with limited visual evidence. Addressing these challenges may involve developing specialized classification decoders for more accurate label identification and incorporating locally perceptive modules to improve the detection of transient actions. These improvements are anticipated to further increase the precision of automated surgical analysis.

\section{Conclusion}

This study introduces OphBiWSSD, a novel framework for dense action localization in ophthalmic surgery that leverages Bidirectional Weight-tied State Space Duality. Evaluation on the OphNet benchmark demonstrates that the proposed architecture achieves state-of-the-art performance, reaching 44.42\% mAP on phases and 43.08\% mAP on operations, representing a substantial improvement over established baselines. The bidirectional selective scan mechanism facilitates the synthesis of non-causal temporal dependencies by integrating preceding and succeeding surgical contexts at each time step. The weight-tied state space duality enforces direction-invariant feature learning via structural parameter sharing, ensuring robust latent state evolution while maintaining computational efficiency. The scalability of the OphBiWSSD framework enables precise, automated recognition of surgical procedures and objective skill assessment in clinical settings.

\bibliographystyle{IEEEtran}
\bibliography{reference}

\end{document}